\documentclass[10pt,twocolumn,letterpaper]{article}

\usepackage{cvpr}              

\usepackage{multirow}
\usepackage{array} 
\usepackage{graphicx}
\usepackage{tabularx}
\usepackage{tikz}
\usetikzlibrary{positioning}

\newcommand\blfootnote[1]{%
  \begingroup
  \renewcommand\thefootnote{}\footnote{#1}%
  \addtocounter{footnote}{-1}%
  \endgroup
}

\definecolor{cvprblue}{rgb}{0.21,0.49,0.74}
\usepackage[pagebackref,breaklinks,colorlinks,allcolors=cvprblue]{hyperref}

\def\paperID{43} 
\def\confName{CVPR}
\def\confYear{2025}

\title{FineCausal: A Causal-Based Framework for \\Interpretable Fine-Grained Action Quality Assessment}

\author{
Ruisheng Han$^1$, Kanglei Zhou$^2$, Amir Atapour-Abarghouei$^1$, Xiaohui Liang$^2$, Hubert P. H. Shum$^1$$^\dagger$ \\
$^1$Durham University \quad $^2$Beihang University\\ 
{\tt\small \{ruisheng.han, amir.atapour-abarghouei, hubert.shum\}@durham.ac.uk,}\\
{\tt\small \{zhoukanglei, liang\_xiaohui\}@buaa.edu.cn}
}

\begin{document}
\maketitle
\begin{abstract}
Action quality assessment (AQA) is critical for evaluating athletic performance, informing training strategies, and ensuring safety in competitive sports. However, existing deep learning approaches often operate as black boxes and are vulnerable to spurious correlations, limiting both their reliability and interpretability. In this paper, we introduce \textbf{FineCausal}, a novel causal-based framework that achieves state-of-the-art performance on the FineDiving-HM dataset. Our approach leverages a Graph Attention Network–based causal intervention module to disentangle human-centric foreground cues from background confounders, and incorporates a temporal causal attention module to capture fine-grained temporal dependencies across action stages. This dual-module strategy enables FineCausal to generate detailed spatio-temporal representations that not only achieve state-of-the-art scoring performance but also provide transparent, interpretable feedback on which features drive the assessment. Despite its strong performance, FineCausal requires extensive expert knowledge to define causal structures and depends on high-quality annotations, challenges that we discuss and address as future research directions. Code is available at \url{https://github.com/Harrison21/FineCausal}.
\end{abstract}    
\section{Introduction}
\label{sec:intro}

\blfootnote{$^\dagger$ Corresponding Author}Action Quality Assessment (AQA) has emerged as a pivotal research area for objectively evaluating the quality of performed actions, offering an alternative to subjective human judgment \cite{zhou2024comprehensivesurveyactionquality}. Since its early development by Gordon \textit{et al.} \cite{gordon1995automated}, AQA has become indispensable in addressing challenges related to bias, reliability, and cost in expert evaluations. Its applications span diverse fields such as skill evaluation \cite{zhang2014relative, parmar2021piano}, medical rehabilitation \cite{capecci2019kimore, fieraru2021aifit, zhou2023video,zhou2024magr}, and sports analysis \cite{xu2019learning, zhang2022learning, nekoui2021eagle, liu2024hierarchical,zhou2024cofinal}. In sports, precise AQA is essential for evaluating athlete performance, designing targeted training programs, and preventing injuries.

Sports videos, unlike general videos \cite{zhou2025adaptive}, are sequential in nature and encapsulate explicit procedural knowledge. For example, in competitive diving, athletes perform a series of rapid and complex movements, ranging from decisive take-off, through intricate somersaults and twists, to the precise entry into the water. Even minor variations in take-off angle, body posture during somersaults, or water entry can significantly influence the final score. However, subtle differences are often difficult to discern accurately by the human eye, which limits the reliability of traditional judgment-based assessments.

Existing AQA methods typically face two major challenges \cite{tang2020uncertainty, xu2022finediving, yu2021group, xu2024fineparser, okamoto2024hierarchical}. First, many approaches either disregard valuable background context by focusing solely on masked foreground regions or, when they attempt to incorporate background information, they rely on simplistic fusion techniques (e.g., sigmoid-based fusion of video and mask features). Such approaches can inadvertently introduce spurious correlations from irrelevant environmental cues, thereby undermining the robustness of the assessment. Second, current methods generally lack interpretability. They fail to provide clear insights into how different stages of an action individually contribute to the overall quality score. This deficiency not only hampers the trustworthiness of the system but also limits its practical utility for athletes and coaches, who require detailed feedback to identify specific areas for improvement. In short, without a fine-grained understanding of the action stages, it becomes difficult to pinpoint which aspects of performance are lacking and how to optimize them effectively.

Motivated by these challenges, we propose \textbf{FineCausal}, a novel causal-based framework that enhances both the performance and interpretability of AQA models. Our approach leverages a causal graph to explicitly model the relationships among original video features, fused features, stage features, and the final action score. Instead of relying on rudimentary fusion techniques, FineCausal employs a Graph Attention Network (GAT)-based causal intervention to dynamically balance the contributions of foreground and background information. In addition, we introduce a temporal causal attention module to capture the semantic and temporal dependencies among different action stages. This temporal module provides fine-grained insights into how each sub-action (e.g., take-off, somersault, twist, and entry) contributes to the overall quality score, thereby offering actionable feedback for performance improvement. Overall, these designs not only improve prediction accuracy but also enhance interpretability by revealing which spatial regions and temporal stages are most influential in determining action quality (see Tab.~\ref{tab:ablation_finecausal}).
Our contributions are:
\begin{enumerate}
    \item We introduce \textbf{FineCausal}, the first causal-based AQA framework that enhances interpretability by explicitly modeling the causal dependencies among video features.
    \item We propose a GAT-based intervention module that adaptively integrates both human-centric foreground cues and valuable background context.
    \item We develop a temporal causal attention module to capture fine-grained, stage-wise relationships across time.
\end{enumerate}

\section{Related Work}
\subsection{Action Quality Assessment}
Action Quality Assessment has evolved considerably over the years. Early work by Pirsiavash \textit{et al.} \cite{pirsiavash2014assessing} treated AQA as a regression problem from action representations to scores, while Parisi \textit{et al.} \cite{parisi2016human} evaluated quality based on the correctness of action matches. Later, Parmar \textit{et al.} \cite{parmar2017learning} demonstrated the effectiveness of leveraging spatio-temporal features for score estimation in competitive sports. Recently, Tang \textit{et al.} \cite{tang2020uncertainty} introduced an uncertainty-aware score distribution learning approach to mitigate the ambiguity in judges' scores, and Yu \textit{et al.} \cite{yu2021group} developed a contrastive regression method that leverages video-level features for accurate ranking and prediction. In parallel, Wang \textit{et al.} \cite{wang2021tsa} proposed TSA-Net, which utilizes outputs from a VOT tracker to generate more informative action representations, while Xu \textit{et al.} \cite{xu2022finediving} introduced an action procedure-aware method alongside a fine-grained sports video dataset to further boost AQA performance. Zhang \textit{et al.} \cite{zhang2023logo} also enriched clip-wise representations by integrating contextual group information through a plug-and-play attention module. More recently, Xu \textit{et al.} \cite{xu2024fineparser} propose a fine-grained spatio-temporal action parser that explicitly parses actions in both space and time to focus on human-centric foreground regions; importantly, they also introduce the FineDiving-HM dataset, which provides fine-grained annotations of human-centric foreground action masks for the FineDiving dataset, promoting the development of real-world AQA systems. Complementing this, Okamoto \textit{et al.} \cite{okamoto2024hierarchical} introduce a hierarchical neuro-symbolic approach that uses neural networks to abstract interpretable symbols from video data and applies rule-based reasoning to assess action quality, ultimately generating detailed visio-linguistic reports that offer transparent feedback on what aspects of performance were good or bad. However, their framework does not explicitly explore the causal relationships that explain \emph{why} certain execution elements lead to success or failure. By contrast, our work explicitly models these causal dependencies, both among visual features and across different sub-action stages, enabling a more comprehensive understanding of how specific cues and temporal phases interact to determine the final outcome.

\subsection{Causal Inference}
In deep learning, confounding factors can lead to the capture of spurious correlations between inputs and outputs. Causal inference techniques provide a theoretical framework to disentangle correlation from causation, thereby enhancing model generalizability, robustness, and interpretability. For instance, Nie \textit{et al.} \cite{nie2023chest} adopt a causal perspective for chest X-ray classification by constructing structural causal models and applying backdoor adjustments to mitigate the influence of spurious correlations. Similarly, Carloni \textit{et al.} \cite{carloni2024crocodile} leverage causal inference via contrastive disentangled learning to foster robustness against domain shifts. In the domain of gaze estimation, Liang \textit{et al.} \cite{liang2024confounded} propose a de-confounded approach that separates gaze-relevant features from irrelevant factors using a dynamic confounder bank strategy, leading to significant cross-domain improvements. Moreover, Liu \textit{et al.} \cite{liu2023cross} introduce cross-modal causal relational reasoning for event-level visual question answering, employing both front-door and back-door interventions alongside spatial-temporal transformers to capture fine-grained visual-linguistic interactions. Collectively, these studies demonstrate that incorporating causal inference not only improves the reliability and fairness of computer vision models but also provides a clearer, more interpretable basis for understanding the underlying mechanisms that drive predictions.

\section{Methodology}
\label{sec:formatting}

\begin{figure*}[h]
    \centering
    \includegraphics[width=0.85\textwidth]{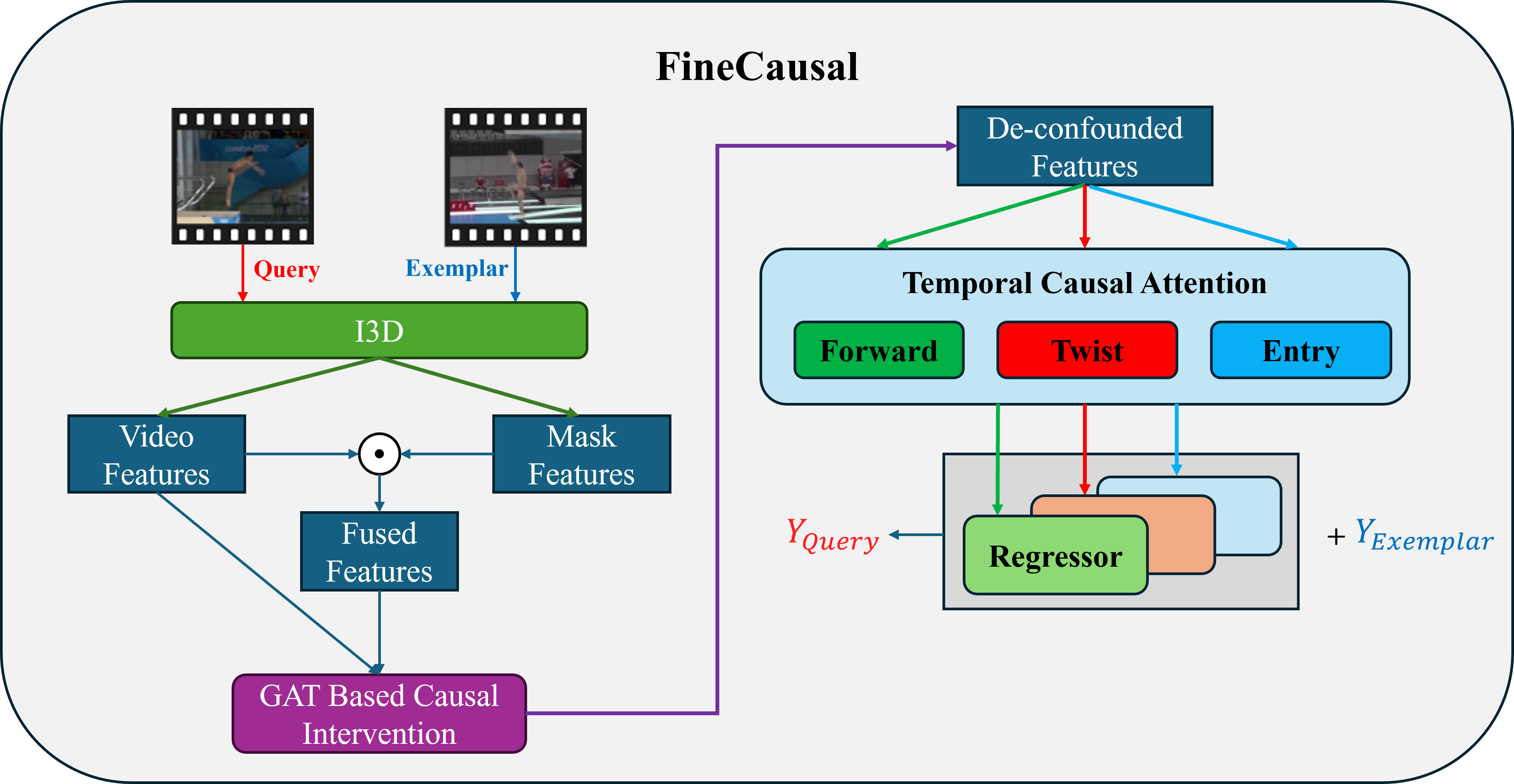}
    \caption{The architecture of FineCausal. The model takes a query and an exemplar video as input, extracting video and mask features through an I3D backbone. These features are fused and refined using a GAT-based causal intervention module to remove spurious correlations, producing deconfounded features. The refined features are then processed through a temporal causal attention mechanism, which decomposes the action into forward, twist, and entry stages. A regressor aggregates stage-wise contributions to predict the query action score \(Y_{\text{Query}}\), adjusted based on the exemplar score \(Y_{\text{Exemplar}}\), ensuring robust AQA.}
    \label{fig:finegait_architecture}
    \vspace{-1em}  
\end{figure*}


In this section, we present our causal-based Action Quality Assessment (AQA) framework, termed \textbf{FineCausal}, as illustrated in Fig.~\ref{fig:finegait_architecture}. Our approach is based on the causal inference methodology proposed in \cite{liang2024confounded}, with the primary goal of effectively utilizing both the foreground (athlete action) and contextual background information (e.g., diving board, audience) to improve prediction accuracy. Unlike prior methods \cite{xu2024fineparser} that apply a straightforward sigmoid-based fusion of video and foreground mask features, which may result in the loss of valuable background information, our proposed GAT-based causal intervention module explicitly identifies and preserves beneficial background features while suppressing spurious correlations.

We first construct a causal graph capturing the relationships among four central variables: \(\mathbf{O}\) (Original video features), \(\mathbf{F}\) (Fused video features), \(\mathbf{S}\) (Stage features), and \(\mathbf{Y}\) (Action score). Next, we elaborate on how our GAT-based intervention mechanism effectively disentangles genuine causal relationships from spurious correlations. As a convention, we use solid arrows to indicate genuine causal effects, while dashed arrows represent spurious causal effects. The following section provides a detailed explanation of the constructed causal graph. 

\subsection{Causal Graph Construction}

\begin{figure}[t]
    \centering
    \includegraphics[width=0.45\textwidth]{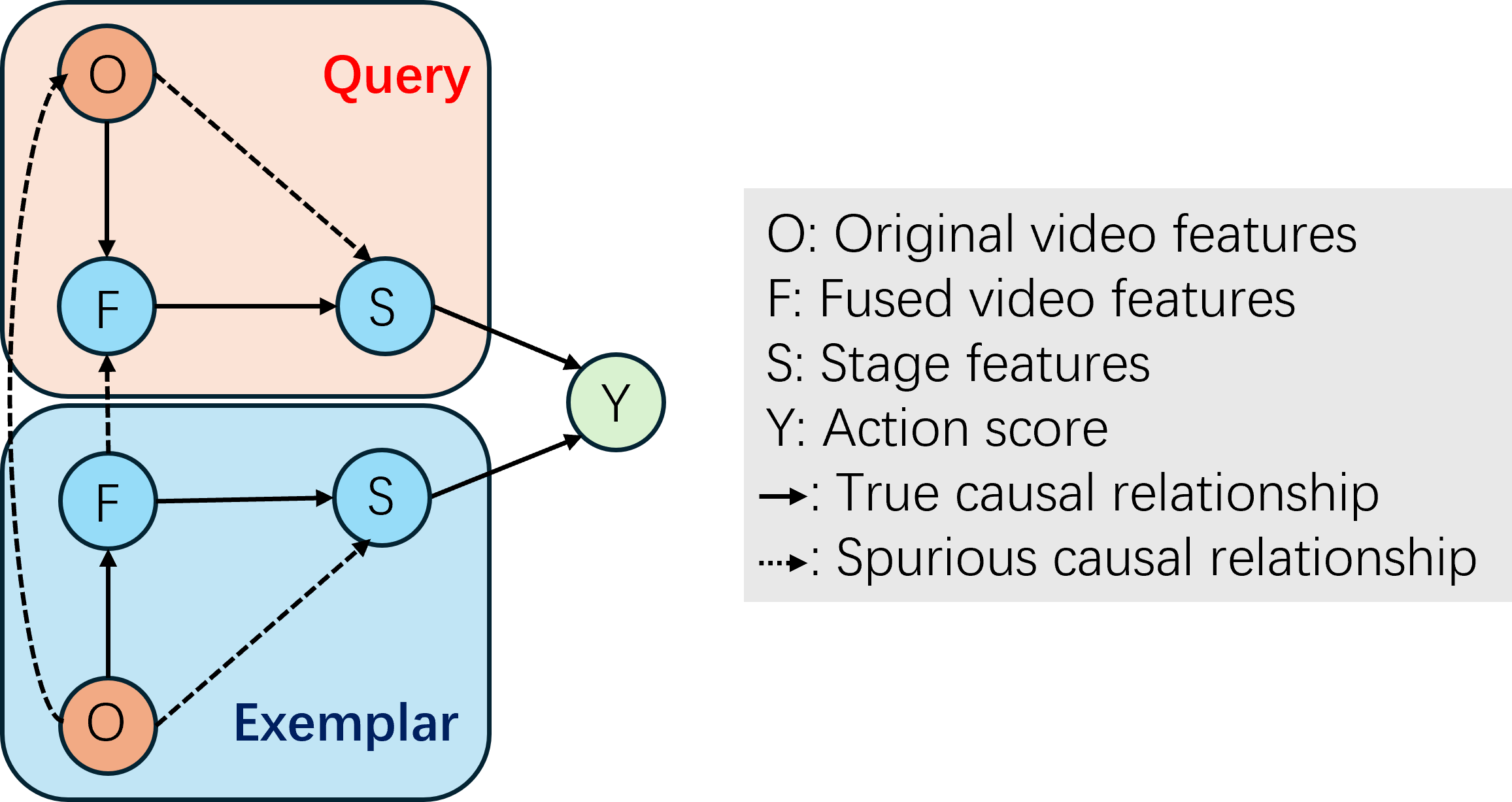}
    \caption{The causal graph of our AQA framework. Nodes represent variables: \(\mathbf{O}\) for original video features, \(\mathbf{F}\) for fused video features, \(\mathbf{S}\) for stage features, and \(\mathbf{Y}\) for final action score. Solid arrows indicate true causal relationships, whereas dashed arrows represent spurious correlations.}
    \label{fig:causal_graph}
    \vspace{-1em}  
\end{figure}
We consider four key variables in the AQA pipeline:
\begin{itemize}
    \item \(\mathbf{O}\): \emph{Original video features} extracted from both the query and exemplar videos (e.g., via a backbone network).
    \item \(\mathbf{F}\): \emph{Fused video features}, obtained by combining \(\mathbf{O}\) with human-centric mask information to emphasize the athlete’s body and reduce irrelevant background.
    \item \(\mathbf{S}\): \emph{Stage features}, encoding the sub-action phases (forward, twist, entry) from \(\mathbf{F}\).
    \item \(\mathbf{Y}\): \emph{Action score}, the final assessment of action quality.
\end{itemize}
Following \cite{liang2024confounded}, we depict these variables in a directed acyclic graph (DAG), illustrated in Fig.~\ref{fig:causal_graph}. The intended causal flow follows:
\[
\mathbf{O} \;\longrightarrow\; \mathbf{F} \;\longrightarrow\; \mathbf{S} \;\longrightarrow\; \mathbf{Y}.
\]
However, prior methods typically rely on a simple sigmoid function to fuse video and mask features. This approach focuses almost exclusively on the masked foreground regions, thereby neglecting background information that could be valuable for AQA. Our hypothesis is that background elements (e.g., diving board structure, audience reaction) also carry informative cues for assessing action quality. While ignoring these cues may limit model performance, their naive incorporation could also introduce misleading dependencies, as the model might inadvertently overfit to environmental biases like lighting variations or audience arrangement. The fundamental causal relationships underpinning AQA are:
\begin{itemize}
    \item \(\mathbf{O} \rightarrow \mathbf{F}\). The original video features \(\mathbf{O}\) (extracted frame-wise) determine the fused features \(\mathbf{F}\). In principle, \(\mathbf{F}\) should highlight the relevant foreground regions (e.g., the diver) and discard irrelevant background.
    \item \(\mathbf{F} \rightarrow \mathbf{S}\). From the fused representation, we segment the action into sub-stages, producing \(\mathbf{S}\). These stage features capture the evolving posture of the athlete during forward, twist, and entry phases.
    \item \(\mathbf{S} \rightarrow \mathbf{Y}\). Finally, each sub-action’s quality affects the overall action score. If a twist phase is poorly executed, it lowers \(\mathbf{S}\) quality and thus reduces \(\mathbf{Y}\).
\end{itemize}
Beyond the intended causal paths, spurious correlations arise due to the influence of shared environmental factors between the query and exemplar videos. These external influences include elements such as background, audience presence, and lighting conditions, which are not directly related to the athlete's performance but can nonetheless impact the final AQA. Key spurious relationships in our framework include:
\begin{itemize}
    \item \(\mathbf{O}_{\text{Exemplar}} \dashrightarrow \mathbf{O}_{\text{Query}}\): The original video features of the exemplar video may influence the original features of the query video due to shared environmental elements, such as the diving board structure or audience positioning, leading to unintended dependencies in feature extraction.
    \item \(\mathbf{F}_{\text{Exemplar}} \dashrightarrow \mathbf{F}_{\text{Query}}\): Similar to the original features, fused video features may also inherit biases from the exemplar video, particularly if the background segmentation is imperfect or if shared contextual elements are mistakenly considered as action-relevant information.
    \item \(\mathbf{O} \dashrightarrow \mathbf{S}\) (for both query and exemplar videos): External factors such as lighting variations or shadows may directly affect the stage-wise decomposition, introducing correlations between the original video features and the extracted stage features that do not genuinely reflect the execution quality.
\end{itemize}
These spurious correlations can mislead contrastive learning-based assessments, where the final action score of the query video is computed based on differences in feature representations between the query and exemplar videos. If these representations are influenced by irrelevant environmental factors, the resulting score may be biased away from the true action quality.

\subsection{Causal Intervention via Graph Attention Networks (GAT)}
\label{sec:causal_intervention}

To mitigate the impact of spurious correlations and enhance the robustness of AQA, we employ Graph Attention Networks (GAT) \cite{velickovic2017graph, xia2023deciphering, Li2025UniEdge, Li2025BPSGCN} to refine the fused video features  through structured feature aggregation. Unlike traditional causal adjustments such as the front-door or back-door criterion \cite{liang2024confounded, nie2023chest}, which require explicit mediators or confounder control variables, the GAT mechanism enables dynamic and adaptive aggregation of node features. In this context, each node represents a fused feature corresponding to a video feature. By computing attention weights conditioned on each node's feature, the model selectively integrates relevant contextual information, such as beneficial background cues, while suppressing less informative or misleading signals.

\subsubsection{Problem Formulation}
In our framework, we model the causal dependencies among the key variables as follows:
\begin{equation}
P(\mathbf{Y} \mid \mathbf{O}, \mathbf{F}, \mathbf{S}) = P(\mathbf{Y} \mid \mathbf{S}) P(\mathbf{S} \mid \mathbf{F}) P(\mathbf{F} \mid \mathbf{O}).
\end{equation}
However, shared environmental elements between query and exemplar videos can introduce unintended correlations in the raw fused features \(\mathbf{F}\) (see Fig.~\ref{fig:causal_graph}). Such spurious correlations may lead to overfitting to background biases, even as potentially valuable contextual information is inadvertently discarded. To address this, we introduce a refined feature representation:
\begin{equation}
\widetilde{\mathbf{F}} = \operatorname{GAT}(\mathbf{O}, \mathbf{F}).
\end{equation}
where the GAT-based intervention is designed to disentangle genuine causal effects from spurious influences. This process ensures that \(\widetilde{\mathbf{F}}\) retains beneficial background cues that contribute to a more accurate assessment of action quality while suppressing irrelevant dependencies induced by shared environmental factors. 

\subsubsection{Implementation via GAT}
Our GAT-based module consists of two key layers:

\noindent\textbf{Initial Feature Propagation}: The first GAT layer computes the refined representation of \(\mathbf{F}\) by attending to feature dependencies across video frames. The updated feature representation is given by:
\begin{equation}
\mathbf{F}^{(1)} = \sigma \left( \sum_{j \in \mathcal{N}(i)} \alpha_{ij} \Theta \mathbf{F}_j \right),
\end{equation}
where \(\Theta\) is a trainable weight matrix, \(\alpha_{ij}\) represents the learned attention weight between nodes \(i\) and \(j\), and \(\mathcal{N}(i)\) denotes the neighborhood of node \(i\) in the feature graph. The attention coefficient \(\alpha_{ij}\) is computed as:
\begin{equation}
\alpha_{i,j} =
\frac{
\exp\left(\mathbf{a}^{\top}\mathrm{LeakyReLU}\left(
\mathbf{\Theta}_{s} \mathbf{x}_i + \mathbf{\Theta}_{t} \mathbf{x}_j
\right)\right)}
{\sum_{k \in \mathcal{N}(i) \cup \{ i \}}
\exp\left(\mathbf{a}^{\top}\mathrm{LeakyReLU}\left(
\mathbf{\Theta}_{s} \mathbf{x}_i + \mathbf{\Theta}_{t} \mathbf{x}_k
\right)\right)},
\end{equation}
where \(\mathbf{\Theta}_s\) and \(\mathbf{\Theta}_t\) are trainable weight matrices for the source and target nodes, respectively, and \(\mathbf{a}\) is a learnable attention vector.

\noindent\textbf{Residual Feature Correction}: The second GAT layer further refines the feature representations by incorporating a residual connection to preserve the original structure of \(\mathbf{F}\):
\begin{equation}
\mathbf{F}^{(2)} = \sigma \left( \sum_{j \in \mathcal{N}(i)} \alpha'_{ij} W' \mathbf{F}_j \right) + \lambda \mathbf{F}^{(1)},
\end{equation}
where \(\lambda\) is a learnable residual weight that balances the contribution of the refined features and the original input.
The final deconfounded representation is then obtained as:
\begin{equation}
\widetilde{\mathbf{F}} = \mathbf{F}^{(2)}.
\end{equation}
By leveraging the GAT-based causal intervention, our framework ensures that \(\widetilde{\mathbf{F}}\) reflect genuine action execution quality. This refined feature representation captures both the critical foreground actions and the useful background context, while suppressing irrelevant dependencies due to environmental biases. As a result, our model achieves improved generalization across diverse scenarios and more robust AQA.

\subsection{Temporal Causal Attention on Stage Features (TCA)}
\label{sec:temporal_causal_attention}

\begin{figure}[t]
\centering
\includegraphics[width=0.45\textwidth]{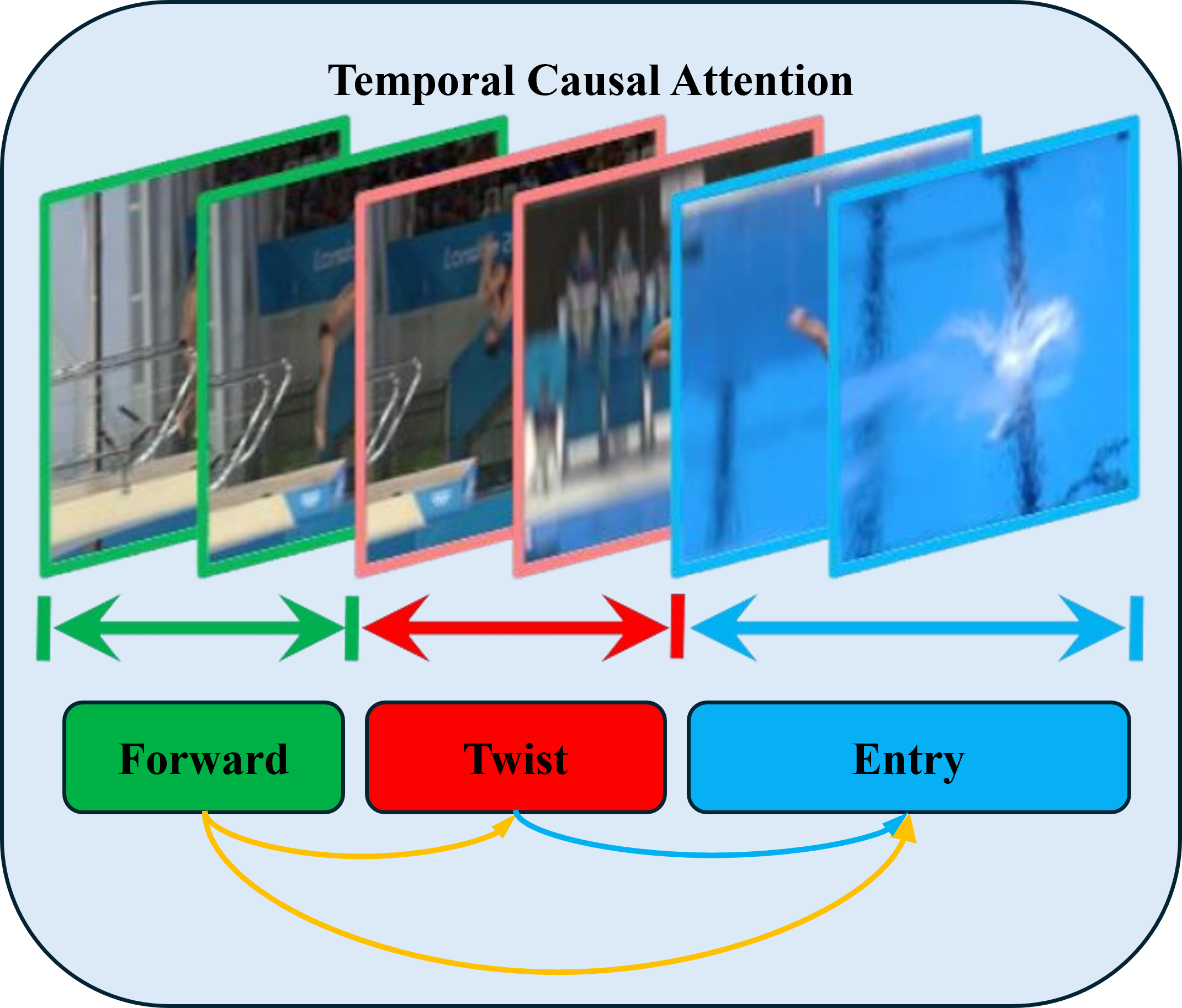}
\caption{Illustration of stage-wise decomposition in action sequences. The movement is split into three stages: forward, twist, and entry. Temporal causal attention models the influence of each stage on the next.}
\label{fig:temporal_causal}
\vspace{-1em}  
\end{figure}

After performing causal intervention on the fused features \(\mathbf{F}\), the features are further decomposed into three distinct stage representations: \(\mathbf{S}_{\text{forward}}\), \(\mathbf{S}_{\text{twist}}\), and \(\mathbf{S}_{\text{entry}}\). These stages represent different phases of an athlete’s movement, such as forward, twist, and water entry. As shown in Fig.~\ref{fig:temporal_causal}, to enhance the interpretability of the model and provide actionable insights for athletes, we introduce \textbf{Temporal Causal Attention} to quantify the influence of one stage on the next. This allows us to determine how much each stage affects subsequent stages, guiding athletes to focus on areas that need improvement.

\subsubsection{Formulation of Temporal Causal Attention}
The causal effect between different stages can be represented as:
\begin{equation}
P(\mathbf{S}_{t+1} \mid \operatorname{do}(\mathbf{S}_t)),
\end{equation}
which measures how an intervention on one stage influences the next. Since these relationships are temporally ordered, we model them using a \textit{ Causal Attention Mechanism} that captures directed dependencies while enforcing a strict forward temporal constraint.

For each stage feature representation \(\mathbf{S}_t\), we apply a masked self-attention mechanism, ensuring that each stage can only attend to itself and previous stages. Given stage features \(\mathbf{S} = [\mathbf{S}_{\text{forward}}, \mathbf{S}_{\text{twist}}, \mathbf{S}_{\text{entry}}]\), the temporal attention scores are computed as:
\begin{equation}
A_{ij} = \frac{\exp( \mathbf{Q}_i \cdot \mathbf{K}_j / \sqrt{d})}{\sum_{k \leq j} \exp( \mathbf{Q}_i \cdot \mathbf{K}_k / \sqrt{d})},
\end{equation}
where \(\mathbf{Q}\) and \(\mathbf{K}\) are query and key representations of the stage features, and the denominator ensures causal masking by summing only over past and current stages.

\subsubsection{Implementation via Transformer-Based Causal Attention}
We implement temporal causal attention using a transformer encoder with masked self-attention layers. The causal attention module restricts information flow to prevent future stages from influencing earlier ones while capturing dependencies across multiple representation subspaces. Additionally, residual connections and layer normalization stabilize training and maintain gradient flow. The forward computation is given by:
\begin{equation}
\mathbf{S}_{t+1}^{(\text{refined})} = \operatorname{softmax} \left( \frac{\mathbf{Q} \mathbf{K}^{\top}}{\sqrt{d}} + \mathbf{M} \right) \mathbf{V},
\end{equation}
where \(\mathbf{M}\) is the causal mask ensuring \( A_{ij} = 0 \) for \( j < i \).

By analyzing the learned attention weights \( A_{ij} \), we can interpret the contribution of each stage to the subsequent one. If a low attention weight is assigned to a transition (e.g., \( A_{\text{twist, entry}} \)), it suggests that poor execution in the twist phase minimally impacts entry, or that the model fails to capture a meaningful connection. This allows us to provide targeted feedback by highlighting which stages significantly influence the final action score, enabling athletes to refine their techniques effectively.

\begin{table}[t]
    \centering
    \renewcommand{\arraystretch}{1.2}

    \begin{tabular}{|>{\centering\arraybackslash}m{2.8cm}|>{\centering\arraybackslash}m{2.1cm}|>{\centering\arraybackslash}m{2.1cm}|}
    \hline
    \multirow{2}{*}{\textbf{Methods}} & \multicolumn{2}{c|}{\textbf{AQA Metrics}} \\
    \cline{2-3}
                                      & $\rho \uparrow$ & $R\text{-}\ell_2 \downarrow$ ($\times 100$) \\
    \hline
    C3D-LSTM~\cite{parmar2017learning}          & 0.6969          & 1.0767 \\
    C3D-AVG~\cite{parmar2019and}            & 0.6801          & 0.6251 \\
    MSCADC~\cite{parmar2019and}              & 0.7688          & 0.9372 \\
    I3D+MLP~\cite{tang2020uncertainty}            & 0.8227          & 0.4878 \\
    USDL~\cite{tang2020uncertainty}                  & 0.8351          & 0.5104 \\
    MUSDL~\cite{tang2020uncertainty}                & 0.8240          & 0.4212 \\
    CoRe~\cite{yu2021group}                  & 0.9308          & 0.3068 \\
    TSA~\cite{xu2022finediving}                    & 0.9361          & 0.2746 \\
    HGCN \cite{zhou2023hierarchical} & 0.9381 & 0.2321 \\
    FineParser~\cite{xu2024fineparser}               & 0.9435 & 0.2602 \\
    FineCausal (Ours)              & \textbf{0.9447} & \textbf{0.2338} \\
    \hline
    \end{tabular}

    \vspace{0.5cm}

    \begin{tabular}{|>{\centering\arraybackslash}m{2.8cm}|>{\centering\arraybackslash}m{2.1cm}|>{\centering\arraybackslash}m{2.1cm}|}
    \hline
    \multirow{2}{*}{\textbf{Methods}} & \multicolumn{2}{c|}{\textbf{TAP Metrics}} \\
    \cline{2-3}
                                      & AIoU@0.5 $\uparrow$ & AIoU@0.75 $\uparrow$ \\
    \hline
    TSA~\cite{xu2022finediving}                    & 0.9239              & 0.5007 \\
    FineParser~\cite{xu2024fineparser}               & \textbf{0.9946}     & \textbf{0.9467} \\
    FineCausal (Ours)               & \underline{0.9937}     & \underline{0.9453} \\
    \hline
    \end{tabular}
    \caption{Comparisons of performance with state-of-the-art AQA methods on the FineDiving-HM Dataset. The best result is highlighted in \textbf{bold}, and the second-best result is \underline{underlined}.}
    \label{tab:aqa_comparison}


\end{table}

\section{Experiments}

\subsection{Datasets}

\textbf{FineDiving-HM.} The FineDiving-HM \cite{xu2024fineparser} dataset comprises 3,000 videos encompassing 52 action types, 29 sub-action types, and 23 difficulty levels. Each video is annotated with fine-grained temporal boundaries and official action scores. To enhance the reliability and interpretability of our model, human-centric action mask annotations are incorporated to distinguish target action regions from the background. FineDiving-HM includes 312,256 annotated mask frames across 3,000 videos, with each mask segmenting the relevant foreground action. The dataset mitigates the challenges associated with frame-level annotation requirements for recognizing human-centric actions in fine-grained spatial and temporal contexts. Among the 312,256 foreground action masks, 248,713 correspond to individual action frames, while 63,543 represent synchronized diving events. These statistics provide valuable insights for athletes and coaches to analyze competition strategies and assess the prevalence of specific actions in diving.

\subsection{Evaluation Metrics}

\textbf{Action Quality Assessment.} 
In line with prior research~\cite{pan2019action, parmar2017learning, parmar2019and, tang2020uncertainty,yu2021group, xu2022finediving, xu2024fineparser,zhou2025adaptive}, we evaluate our model using Spearman’s rank correlation (\(\rho\), where higher is better) and Relative \(\ell_2\) distance (\(R\text{-}{\ell_2}\), where lower is better), which quantify the model’s ability to predict action scores.

\noindent \textbf{Temporal Action Parsing.} To assess the model’s ability to segment action sequences, we utilize the Average Intersection over Union (AIoU) metric~\cite{xu2022finediving, xu2024fineparser}. AIoU measures the alignment between predicted and ground-truth temporal action boundaries, with higher values indicating superior segmentation accuracy.


\subsection{Implementation Details}

The training objective of our model is defined as the sum of three losses, 
\(
L = L_{\text{SAP}} + L_{\text{TAP}} + L_{\text{Reg}},
\)
where \(L_{\text{SAP}}\) is a focal loss applied to optimize the predicted human-centric action masks in the spatial action parser, ensuring accurate foreground extraction; \(L_{\text{TAP}}\) is a binary cross-entropy loss supervising the temporal action parser by comparing the predicted step-transition probabilities with the ground-truth transitions; and \(L_{\text{Reg}}\) is a mean squared error loss that refines the regression in the fine-grained contrastive module by minimizing the discrepancy between the predicted and ground-truth action scores. To further enhance the stability of multitask learning, we incorporate an automated loss weighting mechanism \cite{kendall2018multi} that dynamically adjusts the contribution of each task loss based on its respective uncertainty, thereby preventing any single objective from dominating the training process. The overall training time of our model is approximately one day.
We adhere to the experimental framework established in \cite{xu2024fineparser}, employing the I3D \cite{carreira2017quo} model pre-trained on the Kinetics dataset as the backbone for both the SAP and TAP modules. The learning rates for these modules are set to \(10^{-3}\) for SAP and \(10^{-4}\) for TAP, ensuring distinct parameter optimization. Additionally, the backbone network for shared feature extraction is initialized with a learning rate of \(10^{-3}\). Optimization is performed using the Adam optimizer with weight decay as 0. Following prior work \cite{tang2020uncertainty, xu2022finediving, yu2021group}, we extract 96 frames from each video and divide them into 9 snippets, where each snippet consists of 16 consecutive frames sampled at a stride of 10. The scale factor \(L'\) is set to {3}, while the weighting parameters \(\lambda_l\) are assigned values {3, 5, 2}. For training and evaluation, we adopt the exemplar selection strategy outlined in FineDiving-HM \cite{xu2024fineparser}. Consistent with previous studies \cite{tang2020uncertainty, xu2022finediving, yu2021group}, we allocate 75\% of the dataset for training and 25\% for testing.

\subsection{Comparison with the State-of-the-Arts}

To evaluate the effectiveness of our proposed method, we compare our approach, FineCausal, with state-of-the-art action quality assessment (AQA) models on the FineDiving-HM dataset. The results, presented in Tab.~\ref{tab:aqa_comparison}, demonstrate that FineCausal achieves competitive performance across multiple evaluation metrics.

In terms of AQA metrics, FineCausal attains a Spearman’s rank correlation (\(\rho\)) of \(0.9447\), surpassing the previous state-of-the-art (FineParser) performance (\(0.9435\)) and establishing a new benchmark in the field. Notably, FineCausal achieves the best performance in Relative \(\ell_2\)-distance (\(R_{\ell_2}\)), significantly outperforming FineParser with a relative improvement of \(10.16\%\) (\(0.2338\) vs. \(0.2602\)). Compared to other models, FineCausal demonstrates substantial improvements, reducing the \(R_{\ell_2}\) by \(52.05\%\) over I3D+MLP, \(54.18\%\) over USDL, and \(23.73\%\) over CoRe. These improvements can be attributed to the GAT-based causal intervention, which effectively disentangles spurious correlations from genuine performance cues by selectively aggregating relevant features, including useful background information, and filtering out confounding signals.

For temporal action parsing (TAP), FineCausal attains an AIoU@0.5 score of \(0.9907\) and AIoU@0.75 of \(0.9453\), closely aligning with FineParser (\(0.9946\) and \(0.9467\), respectively). While FineParser slightly outperforms our approach, FineCausal maintains strong performance with only a minor decrease of \(0.39\%\) in AIoU@0.5 and \(0.15\%\) in AIoU@0.75. We attribute this slight drop to the GAT-based causal intervention module: while it effectively disentangles causal features by integrating foreground and background information, the dynamic re-weighting process can introduce minor temporal inconsistencies, making the temporal boundaries of features slightly more vague. Despite this, the overall robustness in predicting temporal boundaries remains high.

Overall, our FineCausal method exhibits superior generalization capabilities, particularly in reducing prediction errors, as evidenced by the lowest \(R_{\ell_2}\) score. This improvement highlights the advantage of leveraging causal attention mechanisms to mitigate spurious correlations, ultimately enhancing model reliability in real-world AQA tasks.

\subsection{Ablation Studies}

\begin{table*}[t]
\centering
\begin{minipage}{0.8\textwidth}  
\centering
\renewcommand{\arraystretch}{1.2}
\setlength{\tabcolsep}{6pt}  

\begin{tabular}{|c|c|c|c|c|c|c|}
    \hline
    \multirow{2}{*}{\textbf{Methods}} 
    & \multicolumn{2}{c|}{\textbf{Modules}} 
    & \multicolumn{2}{c|}{\textbf{AQA Metrics}} 
    & \multicolumn{2}{c|}{\textbf{TAP Metrics}} \\
    \cline{2-7}
    & \textbf{GAT} & \textbf{TCA} 
    & $\rho \uparrow$ & $R\text{-}\ell_2 \downarrow$ ($\times 100$) 
    & \textbf{AIoU@0.5 $\uparrow$} & \textbf{AIoU@0.75 $\uparrow$} \\
    \hline
    A & \checkmark & -- & 0.9425 & 0.2572 & 0.9920 & 0.9069 \\
    B & -- & \checkmark & 0.9382 & 0.2707 & \textbf{0.9973} & \textbf{0.9666} \\
    C & \checkmark & \checkmark & \textbf{0.9447} & \textbf{0.2338} & \underline{0.9937} & \underline{0.9453} \\
    \hline
\end{tabular}

\caption{Merged ablation study results on different modules (GAT and TCA) in FineCausal on FineDiving-HM. Unavailable methods are omitted.}
\label{tab:ablation_finecausal}
\end{minipage}
\vspace{-1em}  
\end{table*}

We conducted an ablation study on the FineDiving-HM dataset to demonstrate the effectiveness of individual parts of \textbf{FineCausal} by designing different modules that selectively incorporate GAT and TCA. Tab.~\ref{tab:ablation_finecausal} summarizes the experimental results.

As shown in Tab.~\ref{tab:ablation_finecausal}, we consider three variants: Method A (GAT only), Method B (TCA only), and Method C (GAT + TCA). Under Spearman’s rank correlation, the AQA performance of the model with GAT (Method A) is 0.9425, while using only TCA (Method B) yields 0.9382. Incorporating both modules (Method C) further improves the correlation to 0.9447. Furthermore, the $R\text{-}\ell_2$ error decreases from 0.2572 in Method A and 0.2707 in Method B to 0.2338 in Method C, indicating that the combined model provides a more precise AQA.

On the temporal action parsing side, significant improvements on AIoU@0.5 and AIoU@0.75 are closely related to the accuracy of the AQA task. Method B (TCA only) achieves the highest AIoU@0.5 of 0.9973 and AIoU@0.75 of 0.9666. Notably, Method C obtains AIoU@0.5 of 0.9937 and AIoU@0.75 of 0.9453, striking a balance between precise temporal segmentation and robust action scoring. These results demonstrate that TCA alone excels in temporal parsing, but when combined with GAT, the model still maintains strong TAP metrics while achieving the best AQA correlation.

In summary, using only GAT or only TCA leads to suboptimal performance in either action quality assessment or temporal action parsing. By integrating both modules, our final FineCausal model achieves the highest Spearman’s rank correlation ($\rho=0.9447$) and the lowest $R\text{-}\ell_2$ error (0.2338), while maintaining competitive TAP accuracy. This confirms the complementary benefits of the GAT and TCA modules for comprehensive AQA.

\begin{figure}[!htbp]
    \centering
    \begin{subfigure}[b]{0.48\textwidth}
        \centering
        \includegraphics[width=\linewidth]{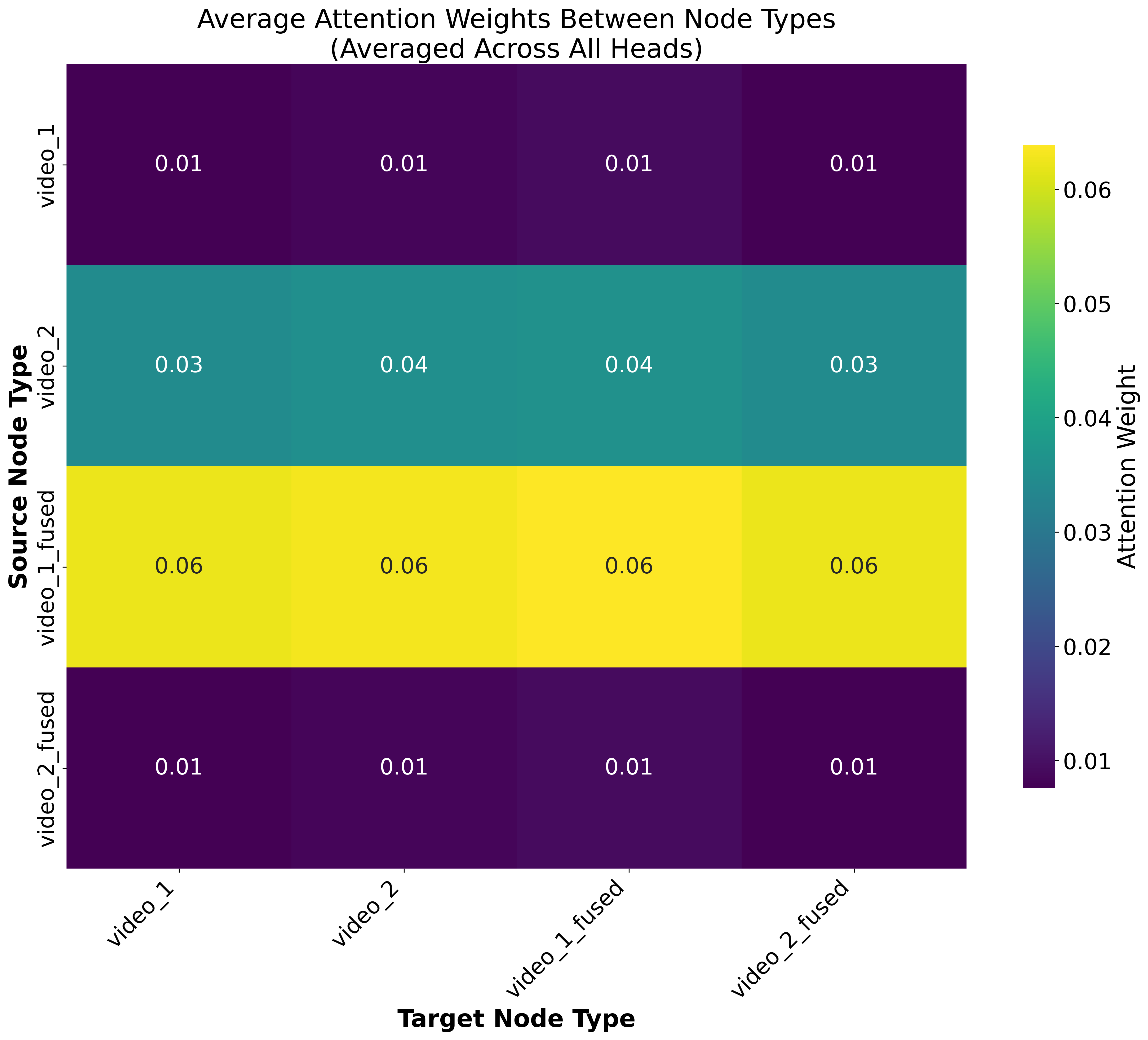}
        \caption{GAT attention map.}
        \label{fig:gat_attention}
    \end{subfigure}
    \hfill
    \begin{subfigure}[b]{0.3\textwidth}
        \centering
        \includegraphics[width=\linewidth]{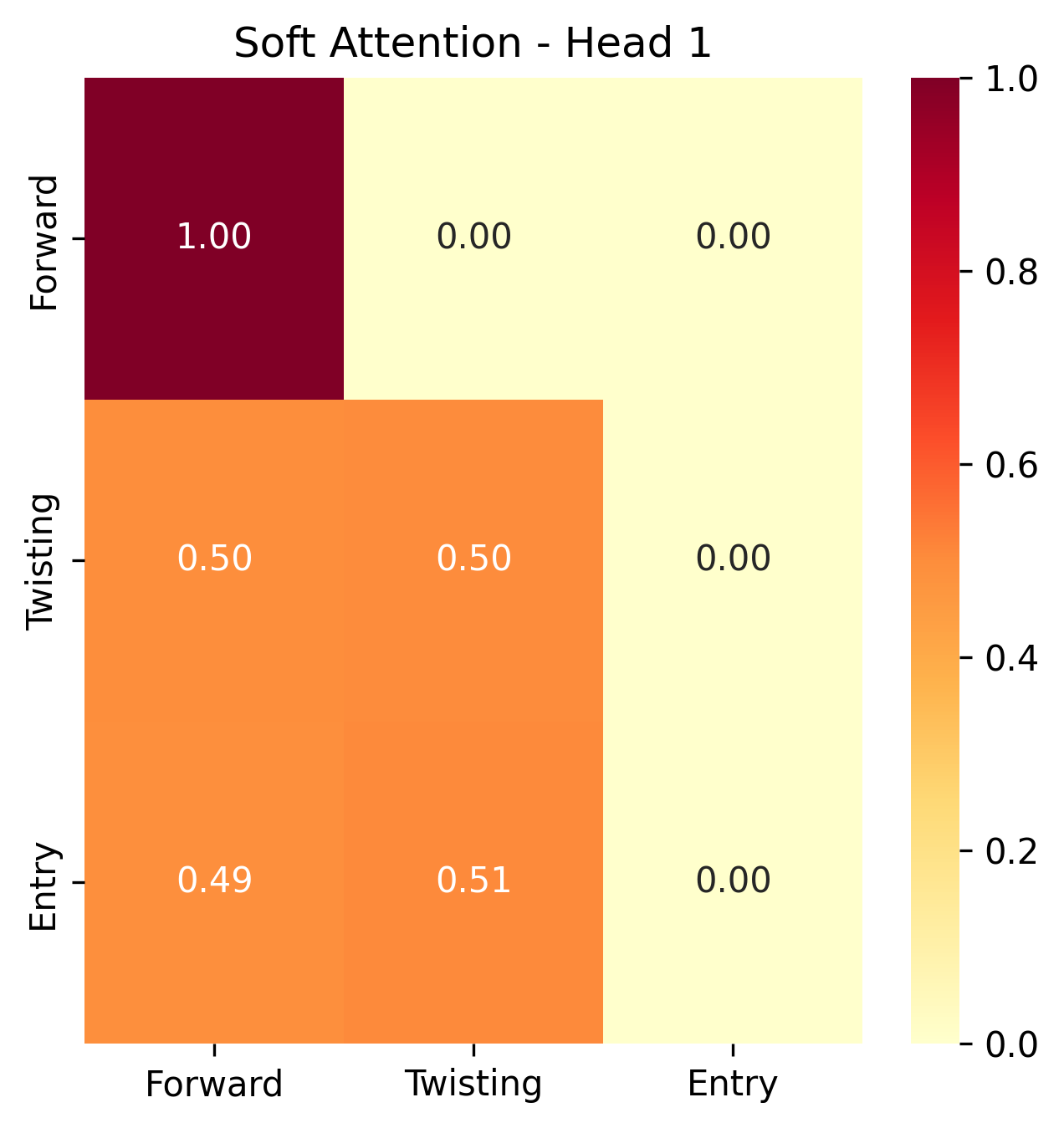}
        \caption{Temporal sub-action attention map.}
        \label{fig:temporal_attention}
    \end{subfigure}
    \vspace{-1em}  
    \caption{Visualization of attention mechanisms in our framework. 
    (a) GAT attention weights between original and fused video features. 
    (b) Temporal attention weights across different sub-action phases.}
    \label{fig:attention_maps}
\end{figure}

\subsection{Visulisation}
\label{sec:visualization}

\noindent To intuitively illustrate how our framework leverages attention mechanisms for robust AQA, we present two sets of attention maps. 

Figure~\ref{fig:gat_attention} shows the GAT attention weights between original and fused video features. Although the fused features of query and exemnplar video (\textit{video\_1\_fused}, \textit{video\_2\_fused}) generally receive higher attention, the original features of query and exemplar video (\textit{video\_1}, \textit{video\_2}) still exhibit non-negligible weights (around 0.01–0.04). This indicates that GAT adaptively integrates both masked (foreground) and raw (background) cues, reinforcing our premise that valuable contextual information in the original video frames should not be discarded. Consequently, the model benefits from a balanced representation that captures the athlete’s motion and relevant environmental context, leading to a more comprehensive AQA.

Figure~\ref{fig:temporal_attention} provides a closer look at the attention distribution across different sub-action phases (e.g., \textit{Forward}, \textit{Twist}, \textit{Entry}). The TSA module selectively highlights important temporal segments, assigning higher weights to sub-actions that significantly impact the overall execution quality. As shown, certain phases (e.g., \textit{Forward}) can dominate the attention map in scenarios where the initial phase is critical to the athlete’s performance. Conversely, when multiple phases (e.g., \textit{Twist} and \textit{Entry}) are equally crucial, the model balances its attention accordingly.

\begin{figure}[t]
    \centering
    \includegraphics[width=0.45\textwidth]{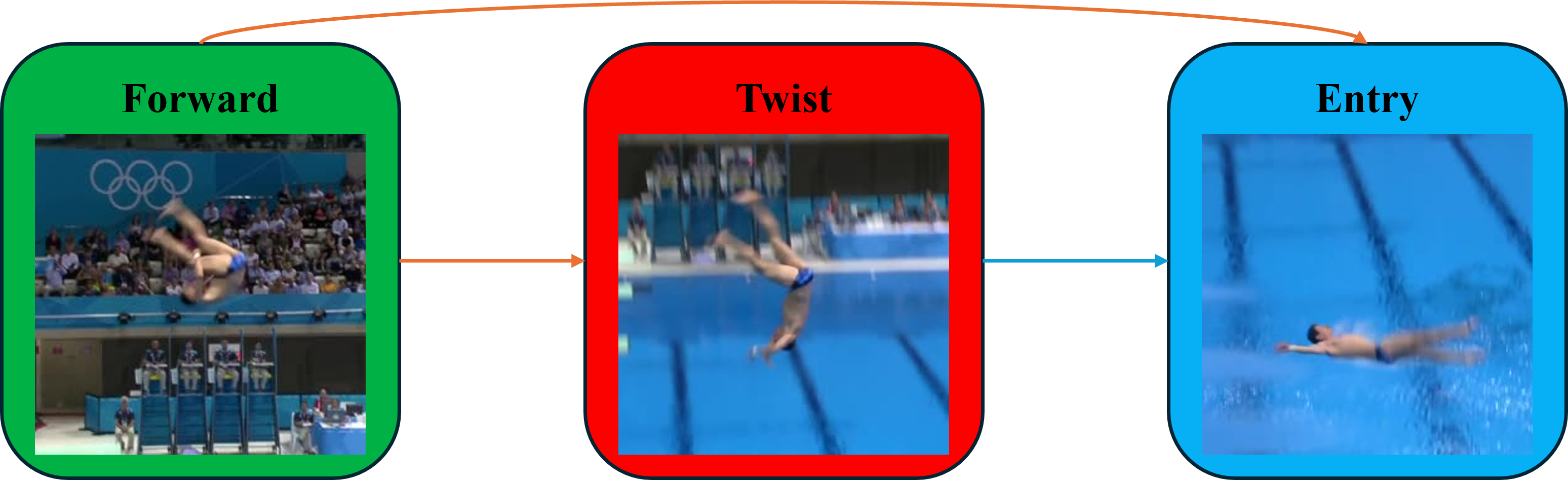}
    \caption{A failure case in which the athlete scores 0 due to an initial mistake in the \textit{Forward} phase, negatively influencing subsequent \textit{Twist} and \textit{Entry} stages. The temporal causal attention highlights how an early misstep can propagate across phases, underscoring the importance of capturing causal dependencies in complex action sequences.}
    \label{fig:failure_case}
    \vspace{-1em}  
\end{figure}

In addition, Figure~\ref{fig:failure_case} illustrates a real-world diving failure where the athlete ultimately receives a zero score. The initial mistake made in the \textit{Forward} stage propagates through the \textit{Twist} and \textit{Entry} stages, leading to a completely unsuccessful dive. The TSA module in our framework is designed to capture such causal dependencies, revealing how early errors can influence later sub-actions. By assigning heightened attention to the stages most affected by the athlete's initial misstep, our model highlights the chain of events that results in the overall failure. This demonstration underscores the importance of robust temporal modeling in AQA, where a single error can have cascading effects on the final outcome.

\section{Conclusion and Discussion}
We presented \textbf{FineCausal}, a novel causal-based framework for AQA that integrates a Graph Attention Network–based causal intervention module and a temporal causal attention module. FineCausal effectively disentangles and balances human-centric foreground features with valuable background context, while the temporal causal attention module captures fine-grained temporal dependencies across action stages. Together, these modules enable the model to learn detailed spatio-temporal representations and provide interpretable feedback on action quality.

Despite achieving state-of-the-art performance and enhanced interpretability, our approach has some macro-level limitations. First, the causal framework often requires substantial expert knowledge to define the underlying causal relationships and design effective interventions, which can be a barrier for applications in domains with less well-established expert guidelines. Second, the high-quality, fine-grained annotations needed, such as those provided in the FineDiving-HM dataset, are critical for the success of causal-based methods, yet such datasets are expensive and time-consuming to create. We believe that addressing these challenges through semi-supervised learning \cite{xin2024maritime} or automated annotation techniques, such as diffusion and Gaussian models \cite{chang23design,wang2025fg,zhao2024gaussianhand}, will be key to broadening the applicability of causality-based AQA systems. Overall, FineCausal establishes a promising baseline for interpretable, causal-based AQA and lays the groundwork for future research in both methodological innovation and dataset development.

\section*{Acknowledgment}
{
This project is supported in part by the EPSRC NortHFutures project (ref: EP/X031012/1).
}

{
    \small
    \bibliographystyle{ieeenat_fullname}
    \bibliography{main}
}


\end{document}